# From Two-Dimensional to Three-Dimensional Environment with Q-Learning: Modeling Autonomous Navigation with Reinforcement Learning and no Libraries

*Ergon Cugler de Moraes Silva*

Getulio Vargas Foundation (FGV)
University of São Paulo (USP)
São Paulo, São Paulo, Brazil

contato@ergoncugler.com
www.ergoncugler.com

**Abstract**

Reinforcement learning (RL) algorithms have become indispensable tools in artificial intelligence, empowering agents to acquire optimal decision-making policies through interactions with their environment and feedback mechanisms. This study explores the performance of RL agents in both two-dimensional (2D) and three-dimensional (3D) environments, aiming to research the dynamics of learning across different spatial dimensions. A key aspect of this investigation is the absence of pre-made libraries for learning, with the algorithm developed exclusively through computational mathematics. The methodological framework centers on RL principles, employing a Q-learning agent class and distinct environment classes tailored to each spatial dimension. The research aims to address the question: **How do reinforcement learning agents adapt and perform in environments of varying spatial dimensions, particularly in 2D and 3D settings?** Through empirical analysis, the study evaluates agents' learning trajectories and adaptation processes, revealing insights into the efficacy of RL algorithms in navigating complex, multi-dimensional spaces. Reflections on the findings prompt considerations for future research, particularly in understanding the dynamics of learning in higher-dimensional environments.

## 1. Introduction

Reinforcement learning (RL) algorithms have emerged as important tools in the realm of artificial intelligence, enabling agents to learn optimal decision-making policies through interaction with their environment and feedback mechanisms (Barto, 1997; Sutton & Barto, 1998; François-Lavet et al., 2018; Henderson et al., 2018; Li, 2018; Zhao et al., 2023). The versatility of RL extends to various spatial contexts, ranging from two-dimensional (2D) planes to three-dimensional (3D) spaces, each presenting unique challenges and opportunities for learning and adaptation (Lin et al., 2020; Kulathunga, 2022).

In this paper, we embark on an exploration of the performance of RL agents operating within both 2D and 3D environments, aiming to elucidate the dynamics of learning across different spatial dimensions. As a central aspect to this paper, no pre-made libraries were used for learning modeling. That is, an algorithm was created exclusively with mathematical modeling, free of libraries at the center of the learning code.

The methodological framework employed in this study is grounded in the principles of RL, emphasizing the implementation of components for training and evaluating agents in diverse spatial settings. Central to this framework is the utilization of a Q-learning agent class, designed to facilitate learning and decision-making processes within the specified environments. This class encapsulates essential functionalities, including the management of Q-values, action selection employing an epsilon-greedy policy, and Q-value updates guided by the principles of Q-learning. Additionally, distinct environment classes tailored to the spatial dimensions are delineated, furnishing methods for environment initialization, reward assignment, and agent movement constraints.

The aim of this research is to investigate the performance of reinforcement learning agents in both two-dimensional (2D) and three-dimensional (3D) environments. By examining the agents' learning trajectories and adaptation processes across different spatial dimensions, we seek to elucidate the dynamics of learning, multi-dimensional settings. So, the research question is: **How do reinforcement learning agents adapt and perform in environments of varying spatial dimensions, particularly in 2D and 3D settings?**

In the 2D scenario, navigation unfolds within a flat plane, facilitated by an Environment class equipped with methods for environment setup, position validation, and reward determination. Conversely, in the 3D context, the exploration extends into a volumetric space, enabled by an Environment3D class that accommodates movement in three dimensions while ensuring adherence to environmental boundaries. Complementing these components is the train_agent function, instrumental in training RL agents over multiple episodes, monitoring rewards and steps per episode to gauge learning progress.

Through an examination of the agents' performance across training episodes, we unravel the trajectory of learning and adaptation within both 2D and 3D environments. In the 2D setting, characterized by a plane with dimensions of 50 units each, the agent demonstrates a discernible refinement in its navigational capabilities over the course of training. Transitioning to the 3D scenario, encompassing a volumetric space with dimensions mirroring the 2D environment, the agent encounters heightened complexity attributable to the additional dimensionality. Nonetheless, the agent's capacity to attain the goal with remarkable efficiency underscores the efficacy of RL algorithms in navigating complex, multi-dimensional spaces.

Subsequently, a succinct exposition on materials and methods delineates the algorithm's intricacies, emphasizing its implementation devoid of learning libraries and reliant solely on computational mathematics. Following this, the presentation of results unfolds through the lens of both two-dimensional (2D) and three-dimensional (3D) perspectives. Lastly, reflections and prospective avenues for future research are expounded upon, illuminating pivotal facets and identifying areas warranting further exploration.

## 2. Materials and methods

The study employed a comprehensive approach to investigate the performance of reinforcement learning agents in both two-dimensional (2D) and three-dimensional (3D) environments. The methodology involved the implementation of key components for training and evaluating the agents' behavior. Central to this was the utilization of a Q-learning agent class, meticulously designed to facilitate learning and decision-making processes within the specified environments. This class encapsulated crucial functionalities, including the management of Q-values, action selection based on an epsilon-greedy policy, and the updating of Q-values following the Q-learning update rule. Moreover, the study delineated distinct environment classes tailored to the spatial dimensions, incorporating methods for environment initialization, reward determination, and agent movement constraints.

For the 2D scenario, an Environment class enabled navigation within a plane, whereas for the 3D context, an Environment3D class facilitated movement in three-dimensional space. Additionally, the study outlined a function, train_agent, instrumental in training the reinforcement learning agent over multiple episodes, tracking rewards and steps per episode to assess learning progress. This methodological framework provided a robust foundation for examining the agents' adaptation and performance within diverse spatial contexts, shedding light on the efficacy of reinforcement learning algorithms in complex environments.

**Table 01. Approaches and codes**

| Approach description | Code description |
|---|---|
| **Both 2D and 3D:**<br><br>This code defines a Q-learning agent class used for reinforcement learning tasks. Upon initialization, the agent is configured with parameters such as the number of possible actions, learning rate, discount factor, and exploration rate (epsilon). The Q-learning agent maintains a Q-table to store Q-values for state-action pairs. The get_q_value method retrieves the Q-value for a given state-action pair. The choose_action method implements an epsilon-greedy policy to select actions, balancing exploration and exploitation. The update_q_value method updates Q-values based on the observed reward and the transition to the next state, following the Q-learning update rule. Overall, this class encapsulates the essential functionalities required for a Q-learning agent to interact with an environment, learn from experiences, and improve its policy over time. | ```python
class QLearningAgent:
    def __init__(self, num_actions, learning_rate=0.1, discount_factor=0.9, epsilon=0.1):
        self.num_actions = num_actions
        self.learning_rate = learning_rate
        self.discount_factor = discount_factor
        self.epsilon = epsilon
        self.q_table = {}

    def get_q_value(self, state, action):
        return self.q_table.get((state, action), 0.0)

    def choose_action(self, state):
        if random.random() < self.epsilon:
            return random.randint(0, self.num_actions - 1)
        else:
            best_action = None
            best_q_value = float('-inf')
            for action in range(self.num_actions):
                q_value = self.get_q_value(state, action)
                if q_value > best_q_value:
                    best_q_value = q_value
                    best_action = action
            return best_action

    def update_q_value(self, state, action, reward, next_state):
        best_next_action =
``` |

| Approach description | Code description |
|---|---|
| | ```
max([self.get_q_value(next_state, next_action)
for next_action in range(self.num_actions)])
        td_target = reward +
self.discount_factor * best_next_action
        td_delta = td_target -
self.get_q_value(state, action)
        self.q_table[(state, action)] =
self.get_q_value(state, action) +
self.learning_rate * td_delta
``` |
| **Both 2D and 3D:**<br><br>This function train_agent is responsible for training a reinforcement learning agent within a given environment over a specified number of episodes, with a limit on the maximum number of steps per episode. During training, it tracks the rewards and the number of steps taken per episode. It iterates over each episode, initializing the state to the environment's starting point and then enters a loop to interact with the environment until the episode ends. Within each episode, the agent selects an action based on its policy, executes the action in the environment, observes the resulting state and reward, and updates its Q-values accordingly using the Q-learning update rule. The function terminates the episode if the agent reaches the goal state or if the maximum number of steps is reached. After each episode, it records the total reward obtained and the number of steps taken. Finally, it returns the lists of rewards and steps per episode, providing insights into the agent's learning progress over the process. | ```
def train_agent(agent, env, num_episodes, max_steps):
    rewards_per_episode = []
    steps_per_episode = []
    for episode in range(num_episodes):
        state = env.start
        total_reward = 0
        num_steps = 0
        while True:
            action = agent.choose_action(state)
            next_state = env.take_action(state, action)
            reward = env.get_reward(next_state)
            agent.update_q_value(state, action, reward, next_state)
            total_reward += reward
            num_steps += 1
            state = next_state
            if next_state == env.goal or num_steps == max_steps:
                rewards_per_episode.append(total_reward)
                steps_per_episode.append(num_steps)
                print(f"Episode {episode + 1}: Total Reward = {total_reward}, Total Steps = {num_steps}")
                break
    return rewards_per_episode, steps_per_episode
``` |
| **2D Approach:**<br><br>This code defines an Environment class representing a 2D environment for a reinforcement learning task. It includes methods to initialize the environment with its dimensions, starting point, and goal location. The is_valid_position method checks whether a given position is within the boundaries of the environment. The get_reward method returns a reward of 1.0 if the agent reaches the goal state; otherwise, it returns a reward of 0.0. The take_action method updates | ```
class Environment:
    def __init__(self, width, height, start, goal):
        self.width = width
        self.height = height
        self.start = start
        self.goal = goal

    def is_valid_position(self, x, y):
        return 0 <= x < self.width and 0 <= y < self.height

    def get_reward(self, state):
        if state == self.goal:
            return 1.0
``` |

| Approach description | Code description |
|---|---|
| the agent's position based on the chosen action, considering movement in four directions: up, down, left, and right, ensuring the new position remains within the boundaries. | ```python
        else:
            return 0.0

    def take_action(self, state, action):
        x, y = state
        if action == 0:    # up
            y += 1
        elif action == 1:  # down
            y -= 1
        elif action == 2:  # left
            x -= 1
        elif action == 3:  # right
            x += 1
        if self.is_valid_position(x, y):
            return (x, y)
        else:
            return state
``` |
| **3D Approach:**<br><br>This code defines an Environment3D class representing a 3D environment for a reinforcement learning task. It initializes the environment with its dimensions, including width, height, and depth, as well as the starting point and goal location. The is_valid_position method checks whether a given position is within the boundaries of the 3D environment. Similar to the 2D version, the get_reward method returns a reward of 1.0 if the agent reaches the goal state; otherwise, it returns a reward of 0.0. The take_action method updates the agent's position based on the chosen action, allowing movement in six directions: up, down, left, right, forward, and backward, ensuring the new position remains within the 3D environment boundaries. | ```python
class Environment3D:
    def __init__(self, width, height, depth, start, goal):
        self.width = width
        self.height = height
        self.depth = depth
        self.start = start
        self.goal = goal

    def is_valid_position(self, x, y, z):
        return 0 <= x < self.width and 0 <= y < self.height and 0 <= z < self.depth

    def get_reward(self, state):
        if state == self.goal:
            return 1.0
        else:
            return 0.0

    def take_action(self, state, action):
        x, y, z = state
        if action == 0:    # up
            y += 1
        elif action == 1:  # down
            y -= 1
        elif action == 2:  # left
            x -= 1
        elif action == 3:  # right
            x += 1
        elif action == 4:  # forward
            z += 1
        elif action == 5:  # backward
            z -= 1
        if self.is_valid_position(x, y, z):
            return (x, y, z)
        else:
            return state
``` |

**Source:** Own elaboration (2024).

# 3. Exploring the results

In this exploration of reinforcement learning, we delve into the dynamics of both two-dimensional (2D) and three-dimensional (3D) environments. Each setting presents unique challenges and learning opportunities for an agent navigating towards a predefined goal. Initially, we scrutinize the parameters and setup of a 2D environment, delineating the dimensions, start and goal positions, and the agent's action space. Subsequently, we transition to a 3D scenario, where we elaborate on the analogous parameters tailored to this spatial dimension. Through an analysis of the agent's performance across training episodes, we unravel the progression of learning and adaptation, witnessing the refinement of decision-making processes and the emergence of efficient navigation strategies.

## 3.1. Two-Dimensional Environment

For the 2D parameters configuration, the width and height represent the dimensions of the 2D environment, set to 50 units each. The start position denotes the initial location of the agent, positioned at coordinates (0, 0), while the goal position specifies the target destination, situated at coordinates (49, 49). The num_actions parameter indicates the number of possible actions the agent can take, which is set to 4, representing directions (up, down, left, right).

The num_episodes parameter determines the number of episodes the agent will undergo during training, defined as 500 episodes. The max_steps parameter sets the maximum number of steps the agent can take within a single episode, limited to 20,000 steps. The learning_rate and discount_factor parameters control the learning behavior of the agent, both set to 0.5, while epsilon defines the exploration rate of the agent's policy, set to 0.2, balancing between exploration and exploitation strategies.

*width = 50*
*height = 50*
*start = (0, 0)*
*goal = (49, 49)*
*num_actions = 4*
*num_episodes = 500*
*max_steps = 20000*
*learning_rate = 0.5*
*discount_factor = 0.5*
*epsilon = 0.2*

The agent's performance reveals a notable refinement throughout the training process. Initially, in episodes like Episode 1, the agent required an extensive number of steps to attain a reward (20,000 steps). As the training advanced, however, the agent exhibited a gradual enhancement in its ability to navigate the environment more effectively. This progression becomes evident in episodes like Episode 10, where the agent achieved the reward in significantly fewer steps compared to its earlier attempts (~8,500 steps).

As training continued, the agent consistently demonstrated improved performance, accomplishing rewards with relatively fewer steps, as observed in Episode 55 and subsequent episodes (~130 steps). This trend persisted throughout the training duration, with the agent consistently achieving rewards with decreasing step counts while maintaining a high success rate. By the concluding episodes, such as Episode 500, the agent consistently attained the reward with notably reduced step counts (precisely 107 steps), indicating its mastery of an efficient policy for navigating the environment. In summary, the agent's advancement over the training period underscores its capacity to learn from experience and optimize its decision-making process to achieve the desired goal more efficiently.

**Figure 01. Cumulative Reward and Steps - 2D**

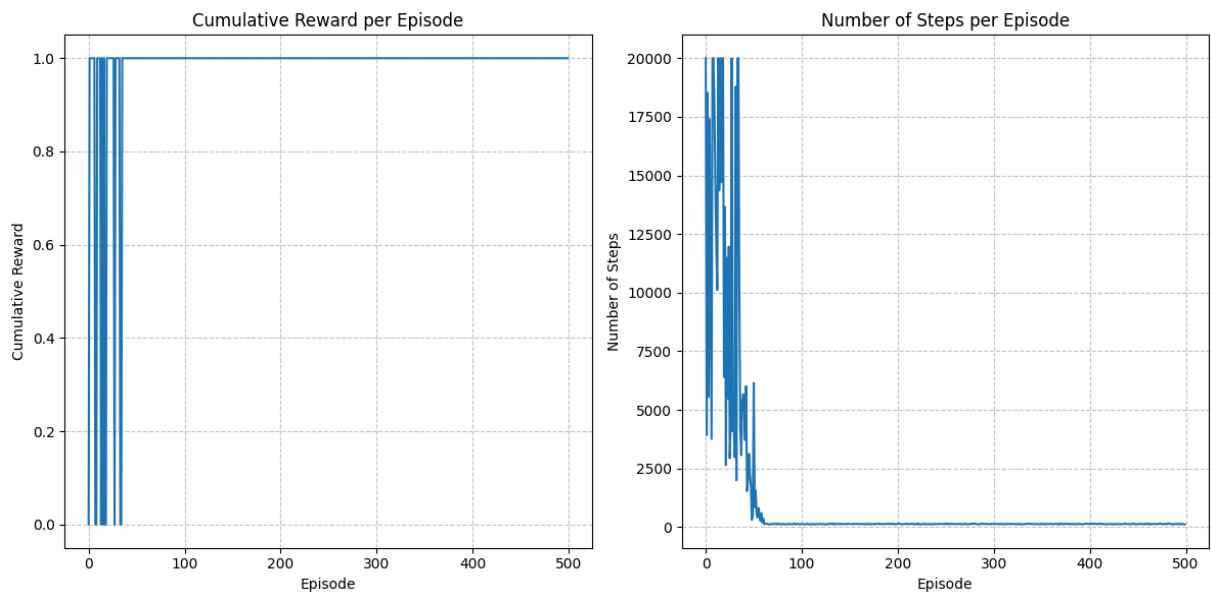

**Source:** Own elaboration (2024).

As noted, there is a stabilization of learning when we reach ~ **Episode 65**, resulting in an agent trajectory referring to the following Figure:

**Figure 02. Path Found - 2D**

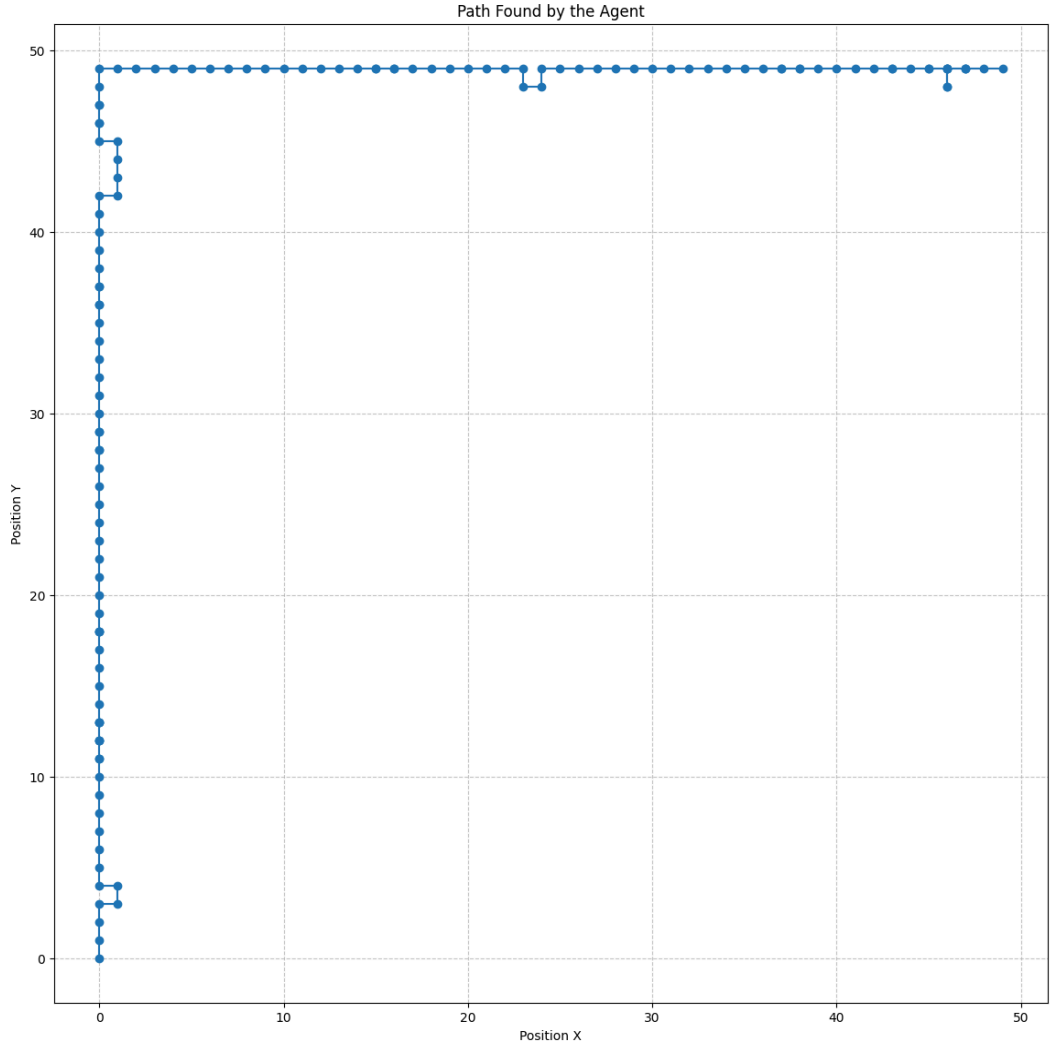

**Source:** Own elaboration (2024).

### 3.2. Three-Dimensional Environment

In the 3D scenario, the parameters are configured as follows: the width, height, and depth specify the dimensions of the 3D environment, each set to 50 units. The start position denotes the initial location of the agent, situated at coordinates (0, 0, 0), while the goal position indicates the target destination, positioned at coordinates (49, 49, 49). The num_actions parameter represents the number of possible actions the agent can take, set to 6, allowing movement in six directions (up, down, left, right, forward, backward).

The num_episodes parameter determines the total number of episodes for agent training, set to 5,000 episodes. The max_steps parameter limits the maximum number of steps the agent can take within an episode, capped at 20,000 steps. Both the learning_rate and discount_factor parameters control the learning behavior, set to 0.5, while epsilon defines the exploration rate of the agent's policy, set to 0.2, maintaining a balance between exploration and exploitation strategies in the learning process.

*width = 50*
*height = 50*
*depth = 50*
*start = (0, 0, 0)*
*goal = (49, 49, 49)*
*num_actions = 6*
*num_episodes = 5000*
*max_steps = 20000*
*learning_rate = 0.5*
*discount_factor = 0.5*
*epsilon = 0.2*

Initially, in Episode 1, the agent required a relatively high number of steps, totaling 20,000, without achieving a reward. However, as the training progressed, the agent's performance improved noticeably. This trend continued throughout the training process, with the agent consistently demonstrating enhanced efficiency in reaching the goal while minimizing the number of steps required.

By Episode ~1,000, the agent consistently achieved rewards with fewer steps, demonstrating its mastery of the task. The latter half of the training period saw the agent achieving rewards with remarkable efficiency, as evidenced by Episode ~1,500, where the agent required only ~200 steps to attain the reward. This trend continued until the final episode, Episode 5,000, where the agent accomplished the task in just 163 steps, highlighting its optimal decision-making capability and mastery of the environment. This exemplifies the effectiveness of the reinforcement learning algorithm in enabling the agent to learn from experience and optimize its behavior to achieve the desired goal efficiently.

**Figure 03. Cumulative Reward and Steps - 3D**

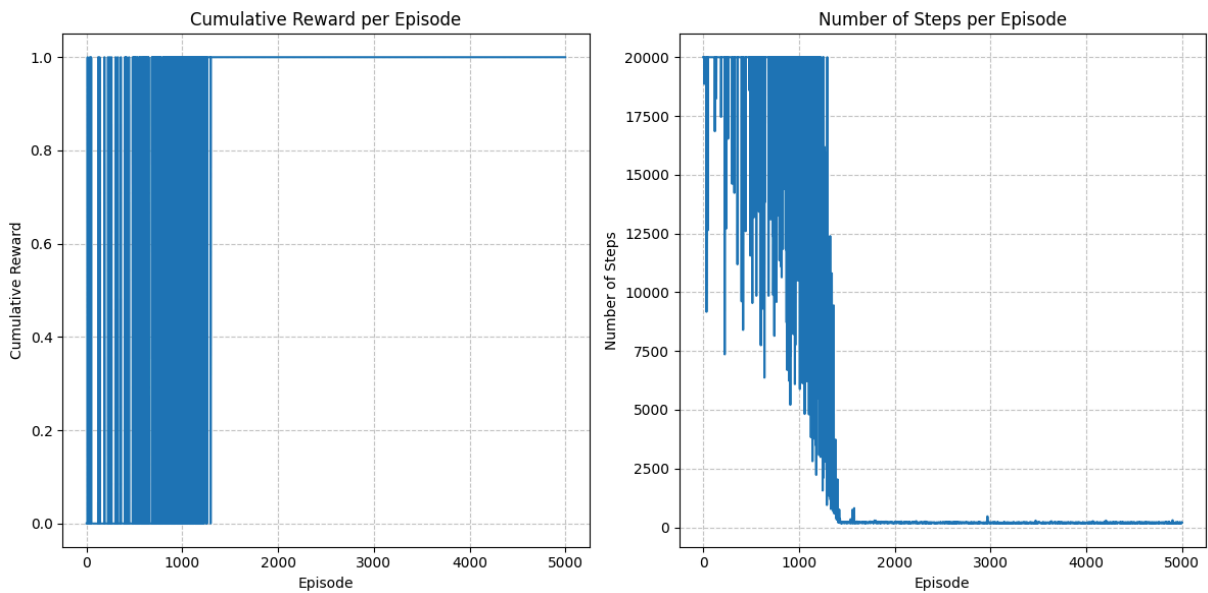

**Source:** Own elaboration (2024).

As noted, there is a stabilization of learning when we reach ~ **Episode 1450**, resulting in an agent trajectory referring to the following Figure:

**Figure 04. Path Found - 3D**

Path Found by the Agent

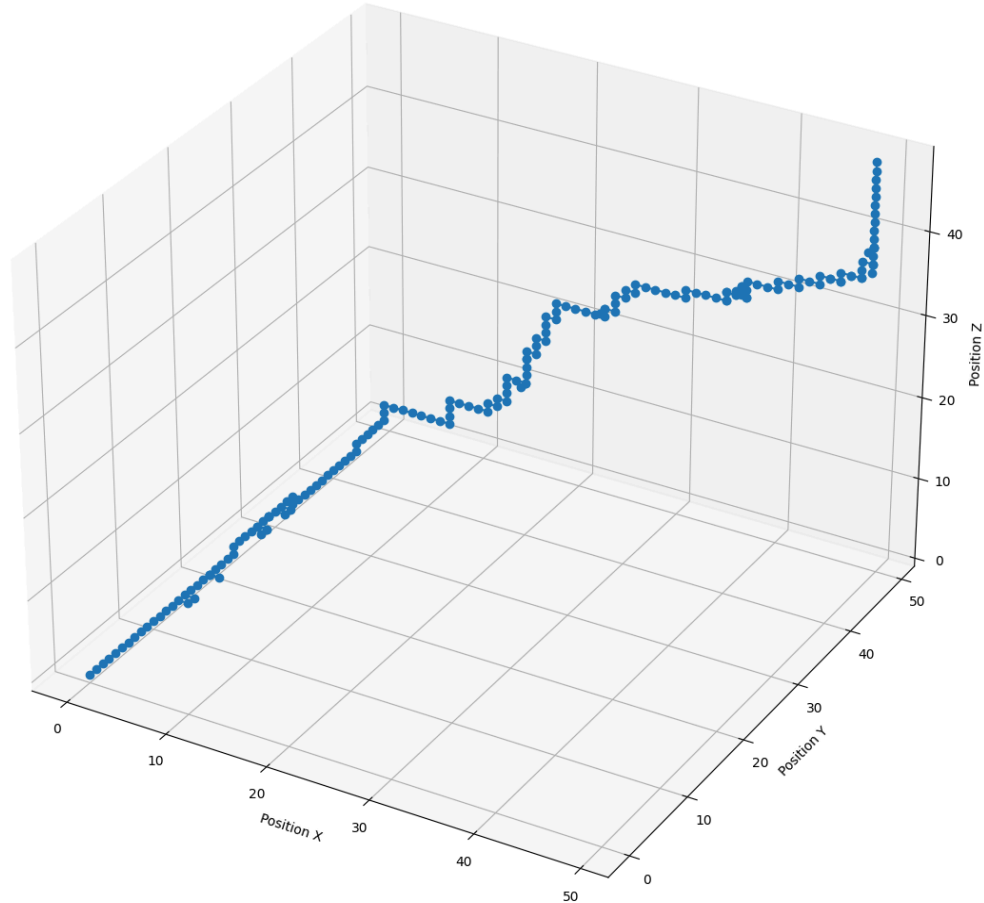

**Source:** Own elaboration (2024).

## 4. Reflections and future work

After conducting a thorough analysis of the Q-Learning algorithm's performance in both 2D and 3D environments, we maintained consistency in parameters, with the primary difference being the presence or absence of the Z-axis.

In the 2D setting, the environment was defined with dimensions of 50 x 50 (Width x Height), starting at position (0, 0) and ending at position (49, 49). The agent had four possible actions, and episodes were limited to a maximum of 20,000 steps. We utilized a learning rate of 0.5, a discount factor of 0.5, and an exploration rate (Epsilon) of 0.2.

For the 3D scenario, we introduced an additional dimension, resulting in a 50 x 50 x 50 (Width x Height x Depth) environment. Despite the consistent start and goal positions, the number of possible actions increased to six to accommodate the additional dimensionality - all other parameters remained unchanged from the 2D scenario.

Upon evaluating the learning curves of the agent in both environments, empirical insights surfaced. Despite the seemingly minor modification in dimensionality, the impact was substantial. Notably, achieving learning stabilization in the 3D environment necessitated a significantly greater number of episodes compared to the 2D counterpart. Specifically, there was an approximate disparity of ~ **65 episodes for the 2D setting**, while the **3D scenario required roughly ~ 1450 episodes until stabilization**. This empirical observation sheds light on the considerable challenge posed by the introduction of an additional dimension and prompts further inquiry into the dynamics of learning in higher-dimensional spaces.

Moreover, it's crucial to note that the difference between 65 episodes and 1450 episodes is approximately 22-fold, indicating that transitioning from the 2D to the 3D environment **demands approximately 22 times more episodes for learning to stabilize**. This stark contrast underscores the substantial increase in computational effort and time required when extending the learning framework from two dimensions to three.

In conclusion, this study does not aim to introduce any novel methodological innovations but rather focuses on the practical application of reinforcement learning (RL) concepts in 2D and 3D environments. Notably, while the study adhered to established RL principles and methodologies, it underscores the computational challenges inherent in scaling RL algorithms to higher-dimensional spaces. Moving forward, further investigation into the dynamics of learning in multi-dimensional environments and the development of strategies to mitigate computational complexities represent promising avenues for future research.

## 5. References

Li, Y. (2018). **Deep Reinforcement Learning:** An Overview. arXiv. Retrieved from https://arxiv.org/abs/1701.07274

Sutton, R. S., & Barto, A. G. (1998). **Reinforcement Learning.** MIT Press. https://mitpress.mit.edu/9780262039246/reinforcement-learning/

Barto, A. G. (1997). **Reinforcement Learning.** In O. Omidvar & D. L. Elliott (Eds.), Neural Systems for Control (pp. 7-30). Academic Press. https://doi.org/10.1016/B978-012526430-3/50003-9

François-Lavet, V., Henderson, P., Islam, R., Bellemare, M. G., & Pineau, J. (2018). **An Introduction to Deep Reinforcement Learning.** Foundations and Trends® in Machine Learning, 11(3-4), 219-354. http://dx.doi.org/10.1561/2200000071

Henderson, P., Islam, R., Bachman, P., Pineau, J., Precup, D., & Meger, D. (2018). **Deep Reinforcement Learning That Matters.** Proceedings of the AAAI Conference on Artificial Intelligence, 32(1). https://doi.org/10.1609/aaai.v32i1.11694

Lin, C., Fan, T., Wang, W., & Nießner, M. (2020). **Modeling 3D Shapes by Reinforcement Learning.** In A. Vedaldi, H. Bischof, T. Brox, & J. M. Frahm (Eds.), Computer Vision – ECCV 2020 (Vol. 12355). Springer. https://doi.org/10.1007/978-3-030-58607-2_32

Zhao, Z., Alzubaidi, L., Zhang, J., Duan, Y., & Gu, Y. (2023). **A comparison review of transfer learning and self-supervised learning:** Definitions, applications, advantages and limitations. Expert Systems with Applications, 122807. https://www.sciencedirect.com/science/article/pii/S0957417423033092

Kulathunga, G. (2022). **A Reinforcement Learning based Path Planning Approach in 3D Environment.** Procedia Computer Science, 212, 152-160. https://doi.org/10.1016/j.procs.2022.10.217

## 6. Author biography

**Ergon Cugler de Moraes Silva** has a Master's degree in Public Administration and Government (FGV), Postgraduate MBA in Data Science & Analytics (USP) and Bachelor's degree in Public Policy Management (USP). He is associated with the Bureaucracy Studies Center (NEB FGV), collaborates with the Interdisciplinary Observatory of Public Policies (OIPP USP), with the Study Group on Technology and Innovations in Public Management (GETIP USP) with the Monitor of Political Debate in the Digital Environment (Monitor USP) and with the Working Group on Strategy, Data and Sovereignty of the Study and Research Group on International Security of the Institute of International Relations of the University of Brasília (GEPSI UnB). He is also a researcher at the Brazilian Institute of Information in Science and Technology (IBICT), where he works for the Federal Government on strategies against disinformation. São Paulo, São Paulo, Brazil. Web site: https://ergoncugler.com/.